\documentclass{article}

\usepackage{microtype}
\usepackage{graphicx}
\usepackage{subfigure}
\usepackage{booktabs} 
\usepackage{multirow}
\usepackage{amsfonts}
\usepackage{hyperref}

\usepackage[accepted]{icml2018}

\icmltitlerunning{Local Contrast Learning}

\begin{document}

\twocolumn[
\icmltitle{Local Contrast Learning}



\icmlsetsymbol{equal}{*}

\begin{icmlauthorlist}
\icmlauthor{Chuanyun Xu}{equal,cqut}
\icmlauthor{Yang Zhang}{cqnu}
\icmlauthor{Xin Feng}{cqut}
\icmlauthor{YongXing Ge}{cqu}
\icmlauthor{Yihao Zhang}{cqut}
\icmlauthor{Jianwu Long}{cqut}

\end{icmlauthorlist}

\icmlaffiliation{cqut}{College of computer science and engineering,Chongqing University of Technology, Chongqing, China}
\icmlaffiliation{cqnu}{College of Computer and Information Science,Chongqing Normal University, Chongqing, China}
\icmlaffiliation{cqu}{School of Software Engineering,Chongqing University, Chongqing, China}

\icmlcorrespondingauthor{Chuanyun Xu}{33677670@QQ.COM}

\icmlkeywords{Machine Learning, Deep Learning,One-shot Learning}

\vskip 0.3in
]



\printAffiliationsAndNotice{\icmlEqualContribution} 

\begin{abstract}
Learning a deep model from small data is yet an opening and challenging problem. We focus on one-shot classification by deep learning approach based on a small quantity of training samples. We proposed a novel deep learning approach named Local Contrast Learning (LCL) based on the key insight about a human cognitive behavior that human recognizes the objects in a specific context by contrasting the objects in the context or in her/his memory. LCL is used to train a deep model that can contrast the recognizing sample with a couple of contrastive samples randomly drawn and shuffled. On one-shot classification task on Omniglot, the deep model based LCL with 122 layers and 1.94 millions of parameters, which was trained on a tiny dataset with only 60 classes and 20 samples per class, achieved the accuracy 97.99\% that outperforms human and state-of-the-art established by Bayesian Program Learning (BPL) trained on 964 classes. LCL is a fundamental idea which can be applied to alleviate parametric model's overfitting resulted by lack of training samples.
\end{abstract}

\section{Introduction}
\label{Introduction}

Deep learning has achieved large successes in many domains of science, business and government for many years because it can provide a kind of end-to-end approach for machine learning and requires very little engineering by hand \cite{lecun_deep_2015}. However, the deep model requires large amount of annotated data for tuning its millions of parameters. Building a large training set for deep learning is sometimes extremely expensive and not acceptable, which hinder deep learning to be applied in more domains. Human can learn a novel concept from just one or few examples \cite{fei-fei_one-shot_2006,lake_human-level_2015}. Can deep model learn a new concept from a small training data, and it don't fall into overfitting? This is an opening and challenging question of machine learning. Most of traditional deep learning approaches try to learn functions that map each high-dimensional sample to lower-dimensional space with the least average training error. In nature, they learn some common and global representations from training data for detecting or classifying patterns in the input\cite{lecun_deep_2015}. It is difficult to draw the representation features from a small training set because the distribution of training data might not be identical with the test set. When the training data are sparse and diverse, it becomes more difficult. This is inconsistent with the human being’s cognitive behaviors that human can more easily distinguish two completely different kinds of objects, such as a car and an apple. So, we can propose a hypothesis that in the lower cognitive level, human recognizes the objects in a specific context by contrasting the objects in the context or in her/his memory instead of drawing universal representations for building a global mapping function.

Based on the key insights, the paper introduced the Local Contrast Learning (LCL) approach for deep learning. LCL makes use of two key ideas from the human cognitive behavior: (a) \textbf{Local context:}Cognition always base on a local context, and only depends on the local context. Local context consists of a set of objects drawn from the objects in global context for cognition. If the context cannot provide enough information for cognition, the new local context must be built by gathering new information or recalling new memories. To human being, the local context  must be a small scenario only with a few target objects. The number of the targets are usually less than seven\cite{miller_magical_1967}. (b) \textbf{Contrast:}In order to identify an object, the recognition is iteratively executed by contrasting the object and each contrastive object in the local context, and the contrast results of the different contrastive objects are compared again among them. The process of the local contrast cognition is shown in Figure 1.

\begin{figure}[ht]
\vskip 0.2in
\begin{center}
\centerline{\includegraphics[height=2in]{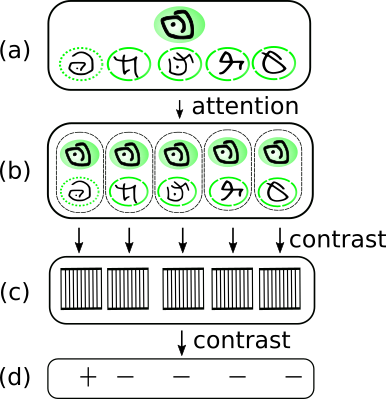}}
\caption{The process of the local contrast cognition. (a) The local contrast context with one recognizing object and five contrastive objects. (b) The contrastive pairs built by moving attention focused on every contrastive object. (c) The difference representations created by contrasting every contrastive pair. (d) Recognition results got by contrasting the difference representations.}
\label{LCL-insight}
\end{center}
\vskip -0.2in
\end{figure}

\section{Local Contrast Learning}
\label{Local Contrast Learning}

LCL is a novel machine learning approach that trains a parametric model by contrasting objects in a local context. The model can learn to distinguish the positive and negative examples in a group of the contrastive samples by contrasting one by one. LCL firstly generates a large amount of contrastive sample groups by randomly sampling classes and exemplars from training set and shuffling them, and then feeds the sample groups into a parametric model to encode the differences, last outputs the activations of the contrastive samples by contrasting the differences. LCL can learn recognization in different local contexts rather than a single global context. 
The workflow of LCL is shown in Figure 2. LCL consists of three major components: Contrast Cognitive Context Constructor (CCCC), Difference Embedding Generator (DEG), and Difference Perceptron (DP). 

\begin{figure}[ht]
\vskip 0.2in
\begin{center}
\centerline{\includegraphics[width=\columnwidth]{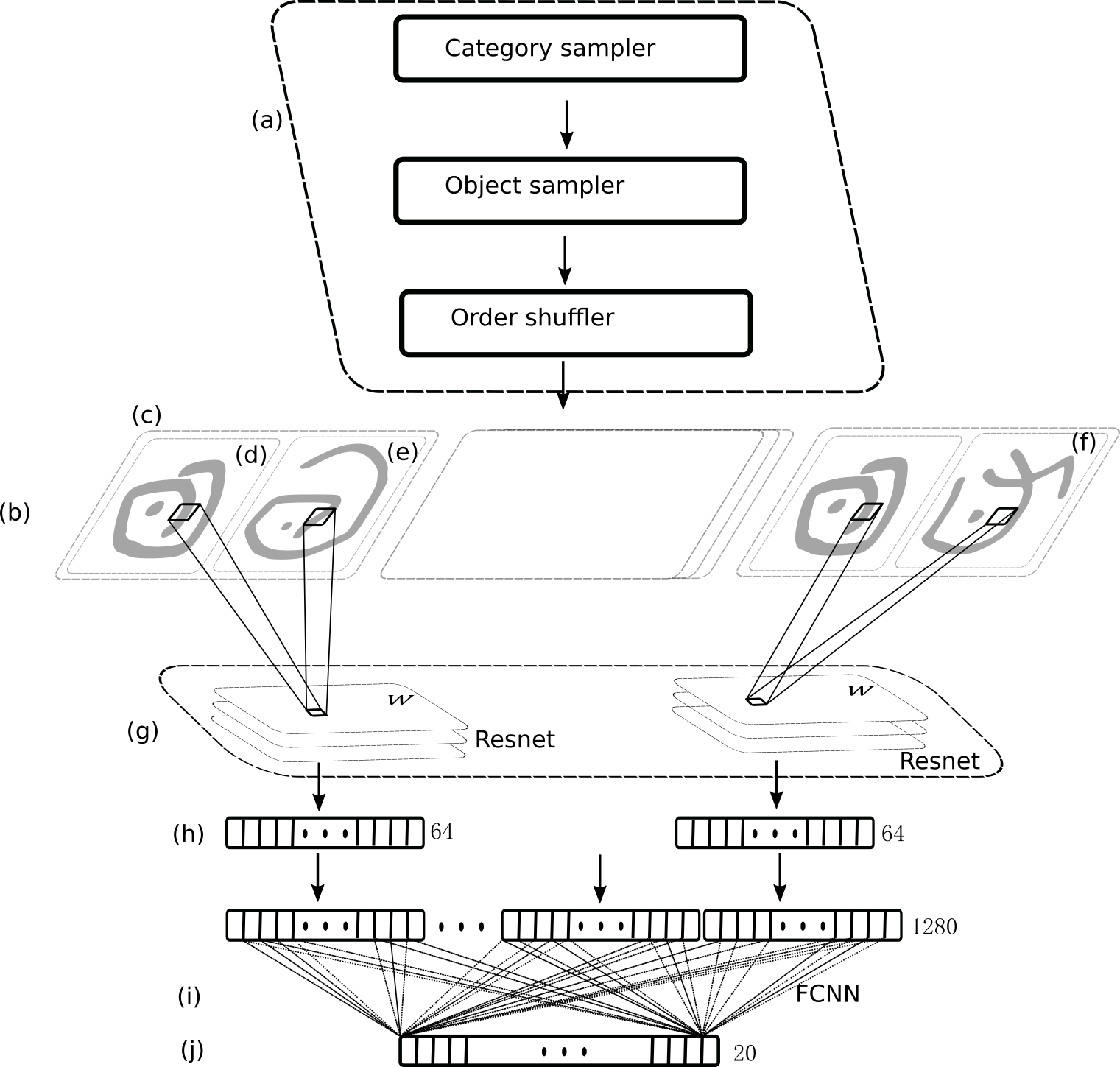}}
\caption{Workflow of Local Contrastive Learning. (a) Contrasting Cognitive Context Constructor including Category Sampler, Object Sampler, order Shuffler, (b) Local Cognitive Context, (c) Contrastive pair, (d) recognizing object, (e) Contrastive Positive Object, (f) Contrastive Negative Object, (g) Difference Embedding Generator that is a parametric model with weights shared, (h) Difference Embedding, (i) Difference Perceptron, (j) Contrastive Pair Local Activation. }
\label{LCL-Workflow}
\end{center}
\vskip -0.2in
\end{figure}

CCCC is responsible for constructing the Local Cognitive Context (LCC) from the labelled support set (training set), denoted as $S=\{S_1,\cdots ,S_c,\cdots,S_{SC}\}$ and $S_c=\{(x_{ci},y_c )\}_{i=1}^K$, where $K$ is the number of the training sample in each category. Here $x_{ci}$ is the $i$th training sample in the $c$th category. Here $y_c \in  \mathbb{C}_S=\left \{c_1,\cdots,c_c,\cdots,c_{SC}\right \}$ is the corresponding label and $SC$ is the numbers of the categories. A LCC consists of one or few recognizing objects from the same category and a set of contrastive objects from different categories, denoted as $LCC=\left((x_c,y_c ),\left \{(x_i,y_i,z_i ) \right \}_{i=1}^L\right)$ is defined by a recognizing object $(x_c,y_c)$, and a set of contrastive objects $\left \{(x_i,y_i,z_i) \right \}_{i=1}^L$. Here $L$ is the numbers of contrastive objects in an LCC. In the contrastive objects, there is one and only one sample whose category is equal to  $y_c$ and that is called contrastive positive object (Corresponding $z_i$ is set to 0), but there is no such a sample $x_i$ that is equal to $x_c$. The contrastive objects except this contrastive positive object are called as contrastive negative object (Corresponding $z_i$ is set to 1). There is only one sample in each category in the contrastive objects.In order to build an LCC, CCCC randomly samples some classes as the recognizing class and contrast classes, then randomly samples one or few instances from each class, and finally randomly shuffles all instances. 

\begin{algorithm}[tb]
   \caption{Generate Local Cognitive Context}
   \label{alg:genLCC}
\begin{algorithmic}
   \STATE {\bfseries Input:} a labelled training set $S$, and the number of contrastive objects $L$
    \STATE 1. Randomly sample $L$ categories from $\mathbb{C}_S$ as contrastive categories.
    \STATE 2. Select anyone as contrastive positive category from the contrastive categories, and others as contrastive negative categories.
    \STATE 3. Iteratively sample one image from each contrastive negative categories as contrastive negative objects.
    \STATE 4. Sample two images from the contrastive positive category, among which one is as recognizing object $(x_c,y_c)$ and another is as contrastive positive object.
    \STATE 5 Concatenate the contrastive positive objects and the contrastive negative objects as contrastive objects $\left \{(x_i,y_i,z_i ) \right \}_{i=1}^L$.
    \STATE 6. Randomly shuffle $\left \{(x_i,y_i,z_i ) \right \}_{i=1}^L$.

\end{algorithmic}
\end{algorithm}

In LCC, the order of contrastive objects is critical. In other word, different LCC can be created from the same contrastive objects by organizing different order. A large quantity of LCC can be created from a small quantity of training samples. For example, in the Small Omniglot dataset with 136 classes, 20 instances each class, in case $L$ equals 20, the 1.18e+44 different LCC can be generated. So, CCCC can provide a large enough number of LCC for training LCL model. It is a critical factor to assure that the training of LCL can stand up to overfitting in facing tiny training set. 

DEG is responsible for contrasting the recognizing object with contrastive objects in LCC one by one and generating a vector representing the differences between them. DEG firstly creates the contrastive pairs $\left\{\left((x_c,y_c ),(x_i,y_i,z_i )\right)\right\}_{i=1}^L$ by grouping the recognizing object and each contrastive object in LCC together, and then the contrastive pairs are iteratively fed into an identical parametric model (such as Resnet\cite{he_deep_2016,he_identity_2016}) in order to get difference embedding that represents the difference of the contrastive pair. 
\[DE=({DE}_1,\cdots,{DE}_i,\cdots,{DE}_L ) \]
\[{DE}_i=deg(((x_c,y_c),(x_i,y_i,z_i)))\]
Here $deg$⁡ is Difference Embedding Generator. 

Theoretically, the embedding model may be any kind of parametric model that can be used as encoder. In the paper, we use Resnet. It must be emphasized that it is a pair of objects inputted into embedding model, among which one is recognizing object and another is contrastive object (positive or negative), and this is like DPSL\cite{grm_deep_2016}, while this is unlike Siamese Network\cite{koch_siamese_2015,kumar_learning_2016}, Matching Nets\cite{vinyals_matching_2016}, or Triplet Network\cite{kumar_learning_2016,hoffer_deep_2015}, Prototypical Networks\cite{snell_prototypical_2017},into which the objects are inputted one by one. This is important for avoiding overfitting. In case of 1000 training samples, if just one object was inputted into embedding model, there were 1000 different inputs, whereas, if two objects were inputted into embedding model, there were 999,000 different inputs. From another view to interpret this, the embedding model maps contrastive object from high-dimensional space to low-dimensional space on condition to recognizing object. It is more difficult to learn the conditional embedding than one object embedding, so this enforce embedding model to sense the difference between recognizing object and contrastive object instead of remembering the object or its features. Moreover, the embedding on condition to recognizing object is local rather than global. The LCL model must learn to embed in different LCC rather than whole training dataset context, so, the model must learn to adapt the tremendous quantity of context instead of single global context. It must be noted that DEG outputs an embedding vector in low dimensional space representing the difference between recognizing object and the contrastive object, instead of a similarity metric or an embedding representation of single object for feeding into similarity function\cite{koch_siamese_2015,kumar_learning_2016,vinyals_matching_2016,hoffer_deep_2015,wang_learning_2014}. It is more robust to map high-dimensional object to vector with 64 dimension than to 1 dimension (In the paper, difference embedding is 64 dimension).

DP is responsible for mapping a group of difference embedding to a vector whose each element represents Contrastive Pair Local Activation (CPLA). 
\[A=(a_i)_{i=1}^L=dp(DE), a_i \in \mathbb{R}\]
Here $dp$⁡ is Difference Perceptron and $a_i$ is the $i$th CPLA. $dp$ is a parametric model, such as fully-connected neural network. It is emphasized that all $DE$ are inputted into $dp$ one time, instead of one bye one. $a_i$ is not determinated only by ${DE}_i$, but by all Difference Embedding. 

CPLA indicates the contrastive object in an LCC is positive or negative. In a LCC, the activation of positive object should be obviously distinguished from the negative object, however, the value of the activation cannot be directly corresponding to the similarity between the recognizing object and the contrastive object. The direct rank for the activations in the entire training set is meaningless, because the contrast results are local in a contrast context. LCL does not aim to train a global similarity metric for contrastive objects. Instead, it just locally identifies the positive object in LCC. It is emphasized that $a_i$ is computed from all difference embedding $DE$ of LCC instead of the difference embedding ${DE}_i$. That means that CPLA is result of contrasting all Difference Embedding’s rather than mapping each Difference Embedding.

\section{Local Contrastive Neural Network}
\label{Local Contrastive Neural Network}
We have provided an implementation of LCL called Local Contrastive Neural Network (LCNN) for the image classification (see Figure 3). LCNN is a multiple layer deep neural network using a Resnet as DEG and a fully-connected layer as DP. The recognizing image and contrastive images in each contrastive pair of LCC are stacked as the input with two channels of Resnet, whose all outputs are concatenated as the input of the fully-connected layer, and its output, CPLA, is a vector whose element is corresponding to each contrastive object. The correspondence between CPLA and contrastive pair is a soft relation which is established by training. In the paper, we make the activation of the negative object tend to one and the activation of the positive object tend to zero. Therefore, the contrastive object with the minimum activation is positive and other are negative. In order to tune the parameters of LCNN, we constructed contrastive loss function \cite{bell_learning_2015,hadsell_dimensionality_2006}. Minimizing the loss function can enlarge the activation gap between the positive object and the negative objects.

\begin{figure}[ht]
\vskip 0.2in
\begin{center}
\centerline{\includegraphics[width=\columnwidth]{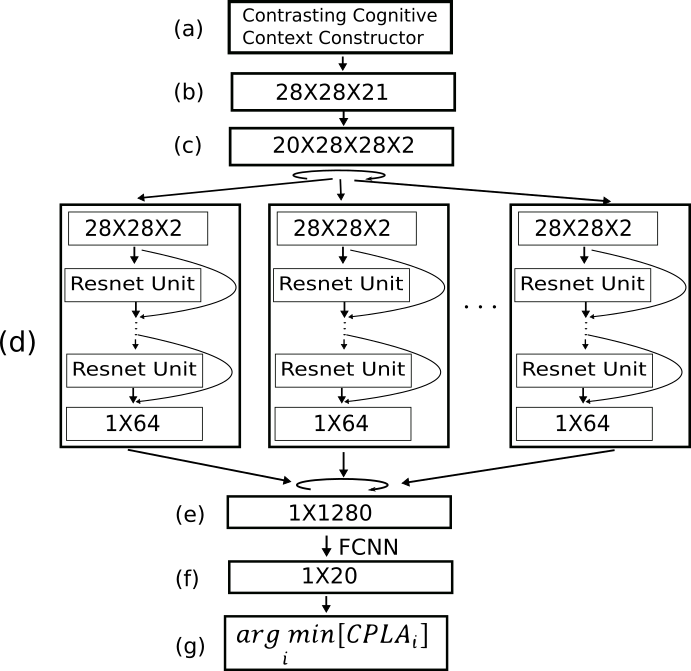}}
\caption{The network architecture of Local Contrastive Neural Network. (a) Contrasting Cognitive Context Constructor. (b) Input included 20 contrastive objects and one recognizing object from Local Contrastive Context where 28 is image size. (c) 20 Contrastive Pairs. (d) Difference Embedding Generator which is a Resnet to be iteratively fed a Contrastive Pair. (e) Concatenating Difference Embedding. (f) Contrastive Pair Local Activation. (g) Minimum Contrastive Pair Local Activation as the positive object. }
\label{LCNN}
\end{center}
\vskip -0.2in
\end{figure}
 
\[L_c(W)=\frac{1}{N}\sum_{1}^{N}\frac{1}{L}\left (\sum_{1}^{L}-\left(z_i\log{a_i}+(1-z_i)\log(1-a_i) \right)  \right ) \]

Where $N$ is number of LCC in a training mini-batch. $L$ is number of the contrastive objects in an LCC. $a_i$ is CPLA and $z_i$ is expected CPLA. $z_i$ equals 0 if the $i$th contrastive objects in LCC is a positive object, otherwise $z_i$ equals 1. $W$ are parameters of LCNN including the parameters $W_{deg}$ of the DEG and the parameters $W_{dp}$ of the DP. 

In order to minimize the loss $L(W)$, the stochastic gradient descent with momentum optimization\cite{rumelhart_learning_1986,sutskever_importance_2013} algorithm is used like Resnet. The momentum of the optimization algorithm is fixed to 0.9. The learning rate starts from 0.1 and after $d_1$ and $d_2$ iterations, the learning rate is respectively set to 0.01, 0.001. $d_1$ and $d_2$ are two hyper parameters. The variance scaling initializer\cite{he_delving_2015} is used for initializing all weight parameters.

One-shot recognition is an extreme case. However, in industrial applications, in order to improve classification accuracy, few-shot recognition is more practicable. The architecture of a few-shot LCNN is shown in Figure 4. The few-shot LCC with a few recognizing samples are built, then they are fed into an identical LCNN for computing CPLA. Finally all CPLA are summed as few-shot CPLA. Similar one-shot recognition, the minimum few-shot CPLA is corresponding the positive object.

\begin{figure}[ht]
\vskip 0.2in
\begin{center}
\centerline{\includegraphics[width=\columnwidth]{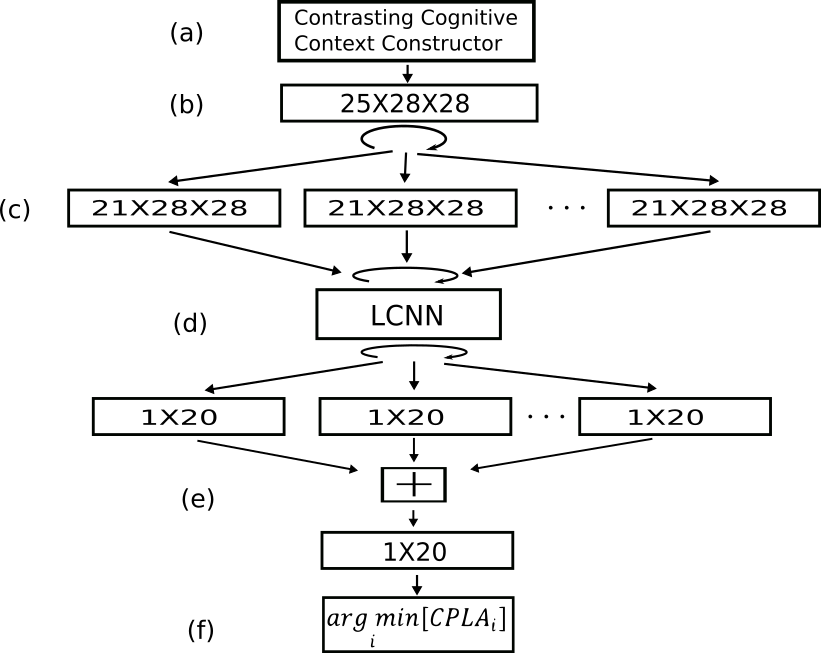}}
\caption{The network architecture of LCNN for few-shot. The figure shows the architecture for 20-way 5-shot classification. (a) Contrasting Cognitive Context Constructor which build LCC with few recognizing objects. (b) The few-shot LCC with 5 recognizing samples. (c) 5 LCC built by unstacking the few-shot LCC. (d) Each LCC is iteratively inputted into a LCNN for generating CPLA. (e) All CPLA are summed as CPLA of the few-shot LCC. (f) The contrastive sample with minimum CPLA is recognized as positive sample. }
\label{fewshotlcnn}
\end{center}
\vskip -0.2in
\end{figure}

\section{Experiments}
\label{Experiments}

In order to evaluate the one-shot (includes few-shot) classification performance of LCNN, we carried out the classification experiments on two benchmark datasets: Omniglot\cite{lake_human-level_2015}, CASIA-HWDB1.1\cite{liu_casia_2011,liu_online_2013}. According to conventional one-shot test protocol\cite{lake_human-level_2015}, all categories of the datasets are divided to two subsets: training set (background set) and test set, and the categories of the training set are disjoint with the categories of the test set. The model will be trained on the categories of the training set and tested on the categories of the test set.

LCNN uses pre-activation Resnet as DEG which are constructed by many pre-activation residual units\cite{he_identity_2016}. Pre-activation residual unit consists of two activation functions (ReLU and BN\cite{ioffe_batch_2015}) and two 3x3 convolutional layers. The first layer of DEG is 3x3 convolutional layer with 16 output channels, followed by $n$ pre-activation residual units with 16 output channels (Here $n$ is a hyper-parameter), $n$ units with 32 output channels,$n$ units with 64 output channels. The last two layers of DEG are activation function (ReLU and BN) and global average pooling \cite{lin_network_2013} that outputs 64 dimension vector. The number of layers of DEG is determined by hyper parameter n and the number of the layers of LCNN equals $(n\times2)\times3+2$.

The learning rate is decayed to 0.01 from 0.1 when the learning steps is larger than $d_1$, and then decayed to 0.001 when the learning steps is larger than $d_2$ (Here $d_1$ and $d_2$ are two hyper-parameters). Maximum training steps $m$ is hyper-parameter. If not specified, in the following experiments, the steps of the learning rate decay $d_1$ and $d_2$ are separately set to 44800 and 51200, and $m$ is 57600. $n$ is set to 20, that is to say, the number of the layers of LCNN is 122. The training mini-batch size is 40. 

Every test result performed on same test set might be different due to three kinds of randomness and computational error, so we run test 100 times on the test set and we report the mean classification accuracy with 95\% confidential interval.

\subsection{Omniglot Based BPL Test Protocol}
Omniglot dataset consists of images across 1623 classes with only 20 images per class, from 50 different alphabets total. The dataset is divided into a background (training or support) set of 30 alphabets with 964 characters and an evaluation (test) set of 20 alphabets with 659 characters. According to the suggestion from Lake et al \cite{lake_human-level_2015-1}, only the background set should be used for training and one-shot learning results are reported using alphabets from the evaluation set. Some one-shot classification results reported ,such as MANN\cite{santoro_meta-learning_2016},Matching Nets\cite{vinyals_matching_2016}, Neural Statistician\cite{edwards_towards_2017}, ConvNet with Memory Module\cite{kaiser_learning_2017}, Prototypical Networks\cite{snell_prototypical_2017}, MAML\cite{finn_model-agnostic_2017}, MetaNet\cite{munkhdalai_meta_2017}, have not accepted the suggestion, so the results don’t match our results in this section and we do comparisons in next section with them.

Except the training set of 30 alphabets with 964 characters, Lake et al\cite{lake_human-level_2015-1} have provided smaller background sets "background small 1" and "background small 2". Each of these contains just 5 alphabets and respectively contains 136 characters and 156 characters. Furthermore, to see how the models perform with more limited training classes, we built two tiny subsets by picking up the first one and first two characters from each alphabet of Omniglot background set, which are respectively called “background tiny1” and “background tiny 2”. The two subsets respectively have 30 characters and 60 characters, and 600 samples and 1200 samples. We trained the LCNN models on all 5 training set for evaluating its performance.

To exactly compare one-shot learning performance, Lake et al\cite{lake_human-level_2015-1,lake_one-shot_2013} developed a standard one-shot test protocol, and we call it BPL test protocol. BPL test protocol provided 20 groups of within-alphabet one-shot classification tasks and each task includes 20 characters selected from an alphabet in the evaluation alphabets. Each character has two examples drawn by two persons, one as test example and another as training example. The test task is, given a test example, to select an example from the 20 training examples, which is from identical character with the test example. So 400 one-shot test trials can be constructed from the 20 test tasks and from which the classification accuracy is calculated. In this section, all experiments are carried out on the 400 trials.

In the experiments, firstly, using CCCC, a mini-batch of LCC are constructed from Omniglot training set, which size is 40 in the all experiments. The number of contrastive objects, $L$, in an LCC is 20, so the performance reported is of 20 way classification. Then the LCC are fed into LCNN for training. The test LCC are created from previous 400 test trials, one trial to one LCC. We augmented the training set through 90, 180 and 270 degrees rotations and resized images to 28 x 28 pixels.

For comparing the performance on small training set with other neural networks approaches, we selected Matching Networks\cite{vinyals_matching_2016} as baseline. We implemented our own basic version without the fully-conditional embedding (FCE). In the basic version, a convolutional network is trained to learn independent embeddings for examples in the training and test set, which consists of a stack of modules, each of which is a 3x3 convolution with 64 filters followed by batch normalization\cite{ioffe_batch_2015}, a Relu non-linearity and 2x2 max-pooling. Maximum training steps is 120,000. The model were trained on 4 dataset: Omniglot Tiny1, Omniglot Tiny2, Omniglot Small1, and Omniglot Small2. The results reported are tested on the previous 400 test trials.

In Table 1, we listed the experiment results along with the prior models, human performance, and Matching Networks10 based on BPL test protocol. 

Using 964 characters for training, LCNN achieved 98.92\% one-shot classification accuracy, and it outperforms the state-of-the-art accuracy 96.7\% established by BPL and also it is better than human’s identification accuracy 95.5\%. The Siamese convolutional network achieved 92.0\% accuracy\cite{koch_siamese_2015} that is much worse than LCNN.

Using Omniglot background small 1 and Omniglot background small 2 for training, LCNN respectively achieved accuracies 98.65\%, 99.17\%   that outperformed BPL that achieved 95.7\%,96.0\% respectively and also outperformed Matching Nets\cite{vinyals_matching_2016} that achieved 47.51\%, 47.26\%. Furthermore, using Omniglot background tiny1 and Omniglot background tiny2, LCNN achieved 92.82\%, 97.99\% one-shot classification accuracies. In this case, even though Matching Nets has just 4 convolutional layers, it got into serious overfitting, which training accuracies are 100.00\% and 99.00\%, however test accuracies are 19.90\% and 37.56\%. LCNN outperformed BPL just using 6 percent samples of BPL.

\begin{table}[t]
\caption{One-shot Classification Accuracy (\%) on Omniglot Based BPL Test Protocol. Results are accuracies averaged over 100 runs and with 95\% confidence intervals where reported. The train set tiny1 and tiny2 consists of first character and first two characters of Omniglot background set.}
\label{sample-table}
\vskip 0.15in
\begin{center}
\begin{small}
\begin{sc}
\begin{tabular}{lrr}
\toprule
Model&\#Chars&Acc.\\Acc.\\
\midrule
Human\cite{lake_human-level_2015-1}&&95.5\\
Pixel kNN\cite{lake_one-shot_2013}&964&21.7\\
Affine model\cite{lake_one-shot_2013}&964&81.8\\
Deep Boltzmann Machines\\\cite{lake_one-shot_2013}&964&62\\
Convolutional Siamese Net\\\cite{koch_siamese_2015}&964&92\\
\hline
Matching Nets\\\cite{vinyals_matching_2016} tiny1&30&19.9\\
Matching Nets\\\cite{vinyals_matching_2016}tiny2&60&37.56\\
Matching Nets\\\cite{vinyals_matching_2016} small1&136&47.51\\
Matching Nets\\\cite{vinyals_matching_2016} small2&156&47.26\\
\hline
BPL small1\cite{lake_one-shot_2013} &136&95.7\\
BPL small2\cite{lake_one-shot_2013}&156&96\\
BPL\cite{lake_one-shot_2013}&964&96.7\\
\hline
LCNN tiny1&30&92.82$\pm$0.08\\
LCNN tiny2&60&97.99$\pm$0.05\\
LCNN small1&136&98.65$\pm$0.06\\
LCNN small2&156&99.17$\pm$0.04\\
LCNN&964&98.92$\pm$0.04\\
\bottomrule
\end{tabular}
\end{sc}
\end{small}
\end{center}
\vskip -0.1in
\end{table}

\subsection{Omniglot Based Variant Test Protocol }

There are a few variant tests proposed by Vinyals, O. et al\cite{vinyals_matching_2016} and Santoro, A. et al\cite{santoro_meta-learning_2016} which have different split about background set and evaluation set and different test trails from BPL Test Protocol\cite{lake_human-level_2015-1}. Vinyals', O.  randomly picked up 1200 characters from Omniglot’s background set and evaluation set as the training set and 423 characters as test set. In Vinyals’ split, the training set and test set have identical alphabets, but have different characters. However, in BPL Test Protocol, the training set and test set have completely different alphabets, that is to say, a model trained on some alphabets is used to classify on some other alphabets. It is more challenging to recognize the images in the novel alphabets than in novel characters in identical alphabets. Some researches also follow Vinyals’, such as such as MANN\cite{santoro_meta-learning_2016},Matching Nets\cite{vinyals_matching_2016}, Neural Statistician\cite{edwards_towards_2017}, ConvNet with Memory Module\cite{kaiser_learning_2017}, Prototypical Networks\cite{snell_prototypical_2017}, MAML\cite{finn_model-agnostic_2017}, MetaNet\cite{munkhdalai_meta_2017}. All test trials are constructed only from the test set and each trial consists of a few characters picked from the test set. Only one image of each character is picked as contrastive object, and one or a few images are picked as recognizing objects.

MetaNet\cite{munkhdalai_meta_2017} has trained and tested on BPL’s split of 30 training alphabets with 964 classes, however it formed 400 trials from the evaluation classes to test the model, so it cannot completely match the BPL’s result which be evaluated on standard 400 trials.

In order to extensively evaluate the performances of LCNN and try to best match other approaches, in this section, we trained and tested the LCCN separately on the test protocol provided by Vinyals, O. et al\cite{vinyals_matching_2016} with 1200 characters and on the test protocol provided by Lake\cite{lake_human-level_2015-1} with 964 characters. Furthermore, in order to show the advantages of LCNN on small data, we trained LCNN on the first 60 characters and the first 156 characters from the 1200 training characters. In order to keep our results reproducible and be available for comparison, unlike other approaches to randomly split on every run, we used same the split on every run (The split will be published). Similar to previous section, we also trained LCNN on "background small 2" and "background tiny 2", but formed trials from the evaluation set rather than the standard 400 trials provided by Lake.

In all experiments in this section, the test trials are created from the test set using Algorithm 1, and the number of test trials $n_{trial}$ is computed by the equation 
$n_{trial}=({EC}\times{KE})/(L+n_{shot})$. Here $EC$ is the number of the characters in the test set. $KE$ is the number of sample per character. $L$ is the number of contrastive objects in a LCC, and $n_{shot}$ is the numbers of the shots.

We carried out 5-way and 20-way one-shot and 5-shot classifications, and the comparisons with published results (as baselines) are shown in Table 2. The models labeled by 60,  156 and 1200 was trained on the test protocol provided by Vinyals, O. , with 60, 156, 1200 training characters and with 423 test characters, in which the training characters and test characters belong to same alphabets. The models labeled by 964, small2, tiny2 was trained on the test protocol provided by Lake respectively with 964,156,60 training characters and with 659 test characters, in which the training characters and test characters belong to completely different alphabets.

\begin{table*}[t]
\caption{One-shot and few-shot Accuracy (\%) on Omniglot Based on Variant Test Protocol. Results are accuracies averaged over 100 times runs with 95\% confidence intervals where reported. ‘-’: not reported.}
\label{Omniglot_Variant}
\vskip 0.15in
\begin{center}
\begin{small}
\begin{sc}
\begin{tabular}{lrrrrr}
\toprule
&&\multicolumn{2}{c}{5-way Acc.}&\multicolumn{2}{c}{20-way Acc.}\\
Model&Chars&1-shot&5-shot&1-shot&5-shot\\
\midrule
Pixels Nearest Neighbor \cite{vinyals_matching_2016} &1200&41.7&-&26.7&-\\
MANN\cite{santoro_meta-learning_2016} &1200&82.8&94.9&&-\\
Matching Nets 10&1200&98.1&98.9&93.8&98.7\\
Neural Statistician\cite{edwards_towards_2017}&1200&98.1&99.5&93.2&98.1\\
ConvNet with Memory Module\\\cite{kaiser_learning_2017}&1200&98.4&99.6&95&98.6\\
Prototypical Networks\cite{snell_prototypical_2017}&1200&98.8&99.7&96&98.9\\
MAML\cite{finn_model-agnostic_2017} &1200&98.7$\pm$0.4&99.9$\pm$0.1&95.8$\pm$0.3&98.9$\pm$0.2\\
MetaNet\cite{munkhdalai_meta_2017} 1200&1200&98.95&-&97&-\\
\hline
LCNN 60&60&98.95$\pm$0.06&99.70$\pm$0.03&96.84$\pm$0.16&99.05$\pm$0.10\\
LCNN 156&156&99.24$\pm$0.05&99.77$\pm$0.03&98.03$\pm$0.13&99.31$\pm$0.09\\
LCNN 1200&1200&98.97$\pm$0.05&99.76$\pm$0.03&98.28$\pm$0.12&99.46$\pm$0.07\\
\hline
MetaNet\cite{munkhdalai_meta_2017} 964&964&98.45&-&95.92&-\\
\hline
LCNN tiny2&60&95.36$\pm$0.09&99.04$\pm$0.05&95.88$\pm$0.20&98.85$\pm$0.11\\
LCNN small2&156&98.54$\pm$0.06&99.63$\pm$0.04&97.01$\pm$0.13&99.15$\pm$0.08\\
LCNN 964 &964&98.86$\pm$0.09&99.70$\pm$0.07&97.77$\pm$0.22&99.33$\pm$0.18\\
\bottomrule
\end{tabular}
\end{sc}
\end{small}
\end{center}
\vskip -0.1in
\end{table*}

\begin{table}[t]
\caption{20-way Classification Accuracy (\%) on HWDB. Results are accuracies averaged over 100 times runs and with 95\% confidence intervals where reported.The models labeled by HWDB60, HWDB200, HWDB3155 was respectively trained on first 60, 200, 3155 characters of HWDB with 20 samples per character.}
\label{Omniglot_Variant}
\vskip 0.15in
\begin{center}
\begin{small}
\begin{sc}
\begin{tabular}{lrrr}
\toprule
Model&Chars&1-shot&5-shot\\
\midrule
Matching Nets\\
\qquad   HWDB60&60&69.78&-\\
\qquad   HWDB200&200&81.67&-\\
\qquad   HWDB3155&3155&85.83&-\\
\hline
LCNN\\
\qquad HWDB60&60&89.85$\pm$0.26&97.70$\pm$0.13\\
\qquad HWDB200&200&96.33$\pm$0.13&99.41$\pm$0.06\\
\qquad HWDB3155&3155&97.32$\pm$0.16&99.60$\pm$0.06\\
\bottomrule
\end{tabular}
\end{sc}
\end{small}
\end{center}
\vskip -0.1in
\end{table}

It must be noted that the experiments in Table 2 are different from the experiments in Table 1 in training set, test set and test trials, so the results in Table 2 cannot match the results in Table 1.

LCNN outperformed prior approaches just using 156 characters, furthermore, and the results using 60 characters is comparable to published results, which used 1200 characters or 964 characters. Comparing the results using 156 characters and the results using 1200 or 964 characters, the latter is not obviously higher than the former, which demonstrate, for training LCNN on Omniglot, 156 character is almost enough, and more data cannot very improve the performance. That proves that LCNN can learn enough knowledge from small data for discriminating objects. Conversely, more training data might imported more noises that confused the models and declined the performance, for instance, the 5-way one-shot accuracy 98.97\% of the model LCNN 964 is slightly lower than the accuracy 99.24\% of the model LCNN 156. 

Comparing the models LCNN tiny2, LCNN small2, LCNN 964, and models LCNN 60, LCNN 156, LCNN 1200, the accuracies of former is lower than latter. That proves that it is more challenging to recognize the images in the novel alphabets than in novel characters in identical alphabets.

\subsection{HWDB}

Like the test on Omniglot, we evaluated LCNN on CASIA-HWDB1.1\cite{liu_casia_2011,liu_online_2013} that is a Chinese handwritten character dataset for machine learning and is more diversified and confusing than Omniglot. CASIA-HWDB1.1 is a widely used dataset that includes 3755 character classes and 300 images per class written by 300 drawers. Therefore, in order to more extensively evaluate the performance and provide a benchmark for other one-shot classification approaches on the handwritten Chinese characters, we defined a one-shot classification of the handwritten Chinese characters. CASIA-HWDB1.1 includes training subset with 240 images per character and  test subset with 60 images per character. However we only use first 20 images per character in the training subset for evaluating the performance of one-shot classification.

We pick up the last 600 characters as test set in 3755 characters (The characters are ordered by GB2312 code), and respectively pick up the first 60, 200, 3155 characters as three different training set: HWDB60, HWDB200, HWDB3155. All images are resized to 64 x 64 and not augmented. In all experiments in this section, the test trials are created using Algorithm 1 on the test set, and the number of test trials $n_{trial}$ is computed by the equation in the previous section. 

For comparing the performance on HWDB with other neural networks approaches, we selected Matching Networks\cite{vinyals_matching_2016} as baseline. The setup of Matching Networks is same with the setup in the previous section, however the maximum training steps is 315,500.

We carried out 20-way one-shot and 5-shot classifications. The results comparing the baselines to LCNN are shown in Table 3. In all experiments, LCNN outperforms the baseline. LCNN with 60 characters achieved accuracy 89.85\% that outperformed the accuracy 85.83\% of Matching Networks with 3155 characters. That demonstrates LCNN can achieve high performance using small data. Comparing the results of LCNN HWDB200 to LCNN HWDB3155, we can find, LCNN HWDB200 achieved almost same one-shot and 5-shot classification accuracy only using 6 percent of the samples of LCNN HWDB3155. That again demonstrates that LCNN can learn enough knowledge from small data for discriminating objects.

\section{Discussion}
Human can identify a novel object by seeing only a few samples per class. That can be explained by the principles of compositionality, causality, learning to learn\cite{lake_human-level_2015-1} and others, and the all principles mainly base on logical reasoning, however before logically thinking, human recognizes objects by intuition based on innate abilities and experiences. Contrast is a kind of innate ability and also is a kind of way enriching experiences. Contrast capability also can be advanced by learning. LCL simulates the capability for recognization. 

The success of LCL is partially attributed to the extremely large quantity of LCC and the three randomness: randomly selecting classes, randomly selecting samples, randomly shuffling contrastive objects. They all enforce LCNN to learn to distinguish objects instead of representing and remembering the training object or its pattern. Learning from contrasting in local context is another crucial factor. Contrasting in local context enforce LCNN to adapt different local context, thus LCNN cannot remember the context information, so, in case of a tiny samples, LCNN would not encounter overfitting. Of course, super deep neural network, such as Resnet, is very important factor, and the its depth may provide enough flexibility for adapting tremendous local context. However, LCL assure the super deep neural network to avoid overfitting in case of tiny samples. Therefore, the super deep neural network contributes to the high classification accuracy, but LCL contributes to its success in training. It is fine design to use Difference Perceptron to contrast the difference embedding instead of learning a similarity metric. That makes the contrast objects in LCC become a list rather than a set, and makes the order of the contrast objects become meaningful. The order is crucial to create tremendous different LCC.

The number of training classes and the number of samples per class are an essential factor to improve the performance. More classes and samples per class can improve the test accuracy. However, the accuracy is only increased when the sample is less. In case the training sample is enough, the increase of sample cannot almost improve the classification accuracy, which is similar with human being’s cognitive behavior that human can learn to recognize object by few samples, but human cannot almost learn more from redundant similar samples. In some cases, the increase of sample might worsen the accuracy, because more samples maybe bring in more noise which can interfere the training. For instance, in Table1, the accuracy 98.92\% of 964 characters is slightly less than the accuracy 99.17\% of 156 characters, because in Omniglot there are some similar characters in different alphabets, that is noise to LCNN.

LCL can be a fundamental tool for setting up artificial intelligence systems. More high level architectures can be invented based on LCL. For example, the hierarchical LCNN could be used to distinguish hierarchical classes. If the objects in LCC are sequential, the sequential LCNN could be created. 

The theoretical explanations for LCL from perspective of cognitive neuroscience and cognitive psychology are expected to study by the researchers in those fields. 

\section{Conclusion}
We proposed a novel deep learning approach named Local Contrast Learning (LCL) for alleviating deep model’s overfitting resulted by lack for training samples. LCL enforce the deep model to adapt tremendous local contexts and carefully capture difference between contrastive object and recognizing object. LCL is able to stably and successfully train deep neural network with more than100 layers using dozens of sample classes and tens of samples each class. The approach achieves state-of-the-art results using 6 percent samples of baselines. The results proves that the deep model can be trained using very small data.

\nocite{langley00}

\bibliography{LCL_paper}
\bibliographystyle{icml2018}

\end{document}